\documentclass{llncs}%

\usepackage[utf8]{inputenc}
\usepackage[T1]{fontenc}
\usepackage[misc]{ifsym}

\usepackage{hyperref}
\usepackage{booktabs}
\usepackage{graphicx}
\usepackage{amsmath}
\usepackage{amssymb}
\usepackage{pifont}
\usepackage[ruled]{algorithm2e}
\usepackage{makecell}

\newcommand{\cmark}{\ding{51}}%

\mainmatter

\begin{document}

\title{DeepAlign: Alignment-based Process Anomaly Correction Using Recurrent Neural Networks}

\author{Timo Nolle \Letter \and Alexander Seeliger \and Nils Thoma \and Max M\"uhlh\"auser}

\institute{
  Technische Universit\"at Darmstadt, Telecooperation Lab, Darmstadt, Germany \\
  \email{\{nolle, seeliger, thoma, max\}@tk.tu-darmstadt.de}
}

\maketitle

\begin{abstract}
In this paper, we propose DeepAlign, a novel approach to multi-perspective process anomaly correction, based on recurrent neural networks and bidirectional beam search.
At the core of the DeepAlign algorithm are two recurrent neural networks trained to predict the next event.
One is reading sequences of process executions from left to right, while the other is reading the sequences from right to left.
By combining the predictive capabilities of both neural networks, we show that it is possible to calculate sequence alignments, which are used to detect and correct anomalies.
DeepAlign utilizes the case-level and event-level attributes to closely model the decisions within a process.
We evaluate the performance of our approach on an elaborate data corpus of 252 realistic synthetic event logs and compare it to three state-of-the-art conformance checking methods.
DeepAlign produces better corrections than the rest of the field reaching an overall $F_1$ score of $0.9572$ across all datasets, whereas the best comparable state-of-the-art method reaches $0.6411$.

\vspace{0.3cm}
\end{abstract}

\section{Introduction}\label{sec:introduction}
Process anomaly detection can be used to automatically detect deviations in process execution data.
This technique infers the process solely based on distributions of the execution data, without relying on an abstract definition of the process itself.
While these approaches can accurately pinpoint an anomaly in a process, they do not provide information about what should have been done instead.
Although, the knowledge about the occurrence of an anomaly is valuable, much more value lies in the knowledge of what was supposed to happen and how to avoid this behavior in the future.

Process mining techniques are centered around the notion of a process model that describes the correct behavior of a process.
Conformance checking techniques can be utilized to analyze process executions for their conformance with a process model.
This method has the benefit of not only detecting deviations from the defined process but also of providing the closest conforming path through the process, thereby correcting it.

The correctness of the conformance checking result depends on the quality of the process model.
Furthermore, a correct execution of a process is not necessarily defined by a correct order of process steps but can depend on a variety of other parameters.
For example, it might not be allowed that the same person executes two consecutive process steps or a process might differ depending on the country it is being executed in.
All these possibilities have to be accounted for both in the process model and the conformance checking algorithm to ensure a correct result.
If no process model is available, conformance checking cannot be used and the creation of a good reference model is a time-consuming task.

An automatic process anomaly correction is therefore desirable, combining the autonomy of an anomaly detection algorithm with the descriptive results from conformance checking.
Against this background, we propose the DeepAlign\footnote{Available on GitHub https://github.com/tnolle/deepalign} algorithm, which combines these two benefits.
It borrows from the field of anomaly detection and employs two recurrent neural networks (RNN), trained on the task of next event prediction, as an approximate process model \cite{tax2018discovery}.
Inspired by the alignment concept from conformance checking, we show that a bidirectional beam search~\cite{sun2017bidirectional} can be used to align a process execution with the process model as approximated by the two RNNs.

DeepAlign can not only detect that process steps have been skipped, but it can also predict which process steps should have been executed instead.
Furthermore, it does not rely on a reference model of the process, nor any prior knowledge about it.
It can be used to automatically detect anomalies and to automatically correct them.

\section{Background}\label{sec:background}
Before we describe the DeepAlign algorithm, we must first introduce some concepts from the field of process mining and deep learning.

\subsection{Process Mining}\label{sec:mining}
Process mining is centered around the idea of human-readable representations of processes called process models.
Process models are widely used in business process management as a tool for defining, documenting, and controlling business processes inside companies.

During the execution of a digital business process, each process step is stored in a database. 
This includes information on when the process step was executed (timestamp), what process step was executed (activity), and to which business case it belongs (case identifier).
These three fundamental bits of event information are the basis for every process mining algorithm and are usually combined into a single data structure called event log.

A log consists of cases, each of which consists of events executed within a process, and some attributes connected to the case (case attributes).
Each event is defined by an activity name and its attributes (e.g., a user who executed the event).
\begin{definition}{Case, Event, and Log.}
  Let $\mathcal{E}$ be the set of all events.
  A case is a sequence of events $c \in \mathcal{E}^{*}$, where $\mathcal{E}^{*}$ is the set of all sequences over $\mathcal{E}$
  Let $\mathcal{C}$ be the set of all cases.
  An event log is a set of cases $\mathcal{L} \subseteq \mathcal{C}$.
\end{definition}

Event logs can be used to automatically discover a process model.
Discovery algorithms analyze the event logs for process patterns and aim to produce a human-readable process model that likely produced the event log.
Multiple discovery algorithms exist, such as the Heuristics Miner~\cite{weijters2011flexible} and the Inductive Visual Miner~\cite{leemans2014process}.

\subsection{Alignments}\label{sec:alignments}
In process analytics, it is desirable to relate the behavior observed in an event log to the behavior defined in a process model.
This discipline is called conformance checking.
The goal of conformance checking is to find an alignment between an event log and a reference process model.
The reference model can be manually designed or be discovered by a process discovery algorithm.

\begin{definition}{Alignment.}
  An alignment~\cite{bose2010trace} is a bidirectional mapping of an event sequence $\sigma_l$ from the event log to a possible execution sequence $\sigma_m$ of the process model.
  It is represented by a sequence of tuples $(s_l, s_m) \in (\mathcal{E}^{\gg} \times \mathcal{E}^{\gg}) \setminus \{(\gg,\gg)\}$, where $\gg$ is an empty move and $\mathcal{E}^{\gg} = \mathcal{E} \cup \{\gg\}$.
  We say that a tuple represents a synchronous move if $s_l \in \mathcal{E}$ and $s_m \in \mathcal{E}$, a model move if $s_l =\ \gg$ and $s_m \in \mathcal{E}$, and a log move if $s_l \in \mathcal{E}$ and $s_m =\ \gg$.
  An alignment is optimal if the number of empty moves is minimal.
\end{definition}

For $\sigma_l = \langle a,b,c,x,e \rangle$ and $\sigma_m = \langle a,b,c,d,e \rangle$,
 the two optimal alignments are
\begin{equation*}
  \begin{array}{|c|c|c|c|c|c|}
    a & b & c & x   & \gg & e \\
    \hline
    a & b & c & \gg & d   & e \\
  \end{array}
  ~~~~\text{and}~~~~
  \begin{array}{|c|c|c|c|c|c|}
    a & b & c & \gg & x   & e \\
    \hline
    a & b & c & d   & \gg & e \\
  \end{array}
\end{equation*}
where the top row corresponds to $\sigma_l$ and the bottom row corresponds to $\sigma_m$, mapping moves in the log to moves in the model and vice versa.

\subsection{Recurrent Neural Network (RNN)}\label{sec:rnn}
Recurrent neural networks (RNN) have been designed to handle sequential data such as sentences.
An RNN is a special kind of neural network that makes use of an internal state (memory) to retain information about already seen words in a sentence.
It is processing a sentence word for word, and with each new word, it will approximate the probability distribution over all possible next words.
Neural networks can be efficiently trained using a gradient descent learning procedure, minimizing the error in the prediction by tuning its internal parameters (weights).
The error can be computed as the difference between the output of the neural network and the desired output.

After the training procedure, the neural network can approximate the probability distribution over all possible next words, given an arbitrary length input sequence.
With slight alterations, RNNs can be applied to event logs, which we will explain further in Sec.~\ref{sec:method}.

\subsection{Beam Search}\label{sec:bs}
In natural language processing, it is common to search for the best continuation of a sentence under a given RNN model.
To find the most probable continuation, every possible combination of words has to be considered which, for a length of $L$ and a vocabulary size of $V$, amounts to $V^L$ possible combinations.
Due to the exponential growth of the search space, this problem is NP-hard.

Instead, a greedy approach can be taken, producing always the most likely next word given a start of a sentence, based on probability under the RNN.
However, this approach does not yield good results because it approximates the total probability of the sentence continuation based only on the probability of the next word.
A more probable sentence might be found when expanding the search to the second most probable next word, or the third, and so on.

Beam search (BS) is a greedy algorithm that finds a trade-off between traversing all possible combinations and only the most probable next word.
For every prediction, the BS algorithm expands only the $K$ most probable sentence continuations (beams).
In the next step, the best $K$ probable continuations over all $K$ beams from the previous step are chosen, and so on.
For $K=1$, BS is equivalent to the greedy 1-best approach explained above.
BS has the advantage of pruning the search space to feasible sizes, while still traversing a sufficient part of the search space to produce a good approximation of the most likely sentence continuation.

The BS algorithm is iteratively applied, inserting new words with each step, until convergence, i.e., the end of a sentence is reached, indicated by the end of sentence symbol.

\subsection{Bidirectional Beam Search}\label{sec:bibs}
The BS algorithm continues a sentence until a special end of sentence symbol is predicted.
However, if the sentence has a defined beginning and end, this approach cannot be used because a unidirectional RNN only knows about the beginning of the sentence and not the end.
This has been demonstrated and been addressed in~\cite{sun2017bidirectional} with a novel bidirectional beam search (BiBS) approach.
Instead of using a single unidirectional RNN, the authors propose to use two separate unidirectional RNNs, one reading the input sentences forwards, and one reading them backwards.

The problem that arises with a gap in the middle of a sentence is that the probability of the resulting sentence, after the insertion of a new word, cannot be computed by a single RNN without re-computation of the remainder of the sentence.
In BiBS, this probability is approximated by the product of the probability of the beginning of the sentence (by the forward RNN), the end of the sentence (by the backward RNN), and the joint probability of inserting the new word (according to both RNNs).
The original BS algorithm is extended to expand the search space based on this joint probability, ensuring a proper fit both for the beginning and the end of the sentence.

The BiBS algorithm is iteratively applied to the original sentence, updating it with each step, until convergence, i.e., no insertions would yield a higher probability in any of the $K$ beams.

\section{DeepAlign}\label{sec:method}
\begin{figure}[t]
  \centering
  \includegraphics[width=0.8\linewidth]{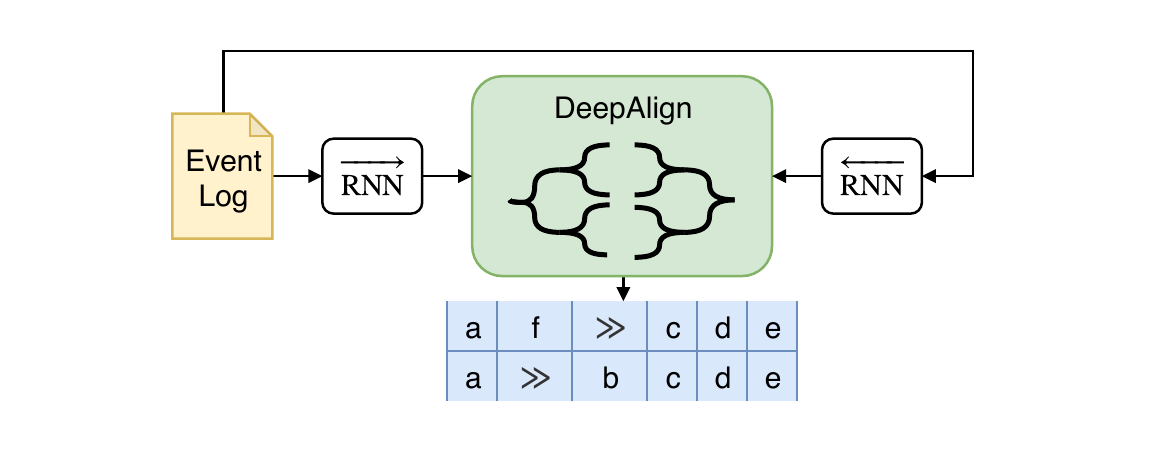}
  \caption{The DeepAlign algorithm makes use of two next event prediction RNNs and an extended bidirectional beam search (green) to produce alignments}
  \label{fig:algorithm}
\end{figure}

In this section we describe the DeepAlign algorithm and all its components.
An overview of the algorithm is shown in Fig.~\ref{fig:algorithm}.
Two neural networks are trained to predict the next event, one reading cases from left to right (forwards), the other reading them from right to left (backwards).
An extended BiBS is then used to transform the input case to the most probable case under the two RNN models. 
Lastly, an alignment is calculated based on the search history of the algorithm.

\subsection{Next Event Prediction}
Next event prediction aims to accurately model the decisions being made in a process.
These decisions are based on multiple parameters, such as the history of a case, the attributes connected to past events, and the case level attributes.
To succeed, a machine learning model must take into account all of these parameters.

\begin{figure}[t]
  \centering
  \includegraphics[width=0.5\linewidth]{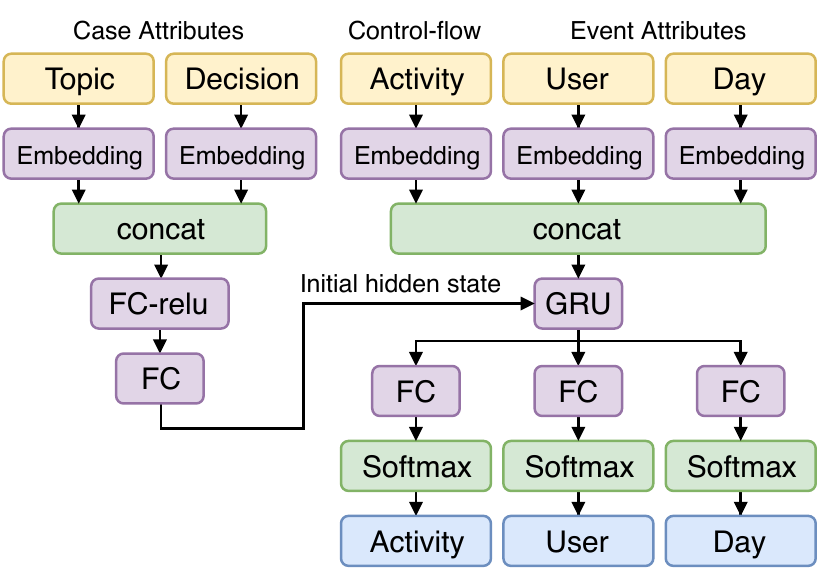}
  \caption{RNN architecture for an event log with two case attributes (\textit{Topic} and \textit{Decision}) and two event attributes (\textit{User} and \textit{Day})}
  \label{fig:architecture}
\end{figure}

In this paper, we propose a new neural architecture for next event prediction.
It has been designed to model the sequence of activities (control-flow), the attributes connected to these activities (event attributes), and the global attributes connected to the case (case attributes).
Figure~\ref{fig:architecture} shows the architecture in detail.

At the heart of the network is a Gated Recurrent Unit (GRU) \cite{cho2014learning}, a type of RNN.
This GRU is iteratively fed an event, consisting of its activity and its event attributes, and must predict the corresponding next event.
Each categorical attribute is fed through an embedding layer to map the values into a lower-dimensional embedding space.
To include the case attributes, we make use of the internal state of the GRU.
Instead of initializing the state with zeros (the default), we initialize it based on a representation of the case attributes.
All case attributes are transformed by a case attribute network, consisting of two fully-connected layers (FC), to produce a real-valued representation of the case attributes.
In other words, we initialize the next event prediction with a representation of the case attributes, thereby conditioning it to predict events according to these case attributes.
Finally, the GRU output is fed into separate FC layers with Softmax activations to produce a probability distribution over all possible attributes of the next event (i.e., the prediction of the next event).

We train the networks with a GRU size equal to two times the maximum case length on mini-batches of size $100$ for $50$ epochs using the Adam optimizer with standard parameters~\cite{kingma2014adam}.
The first layer of the case attribute network has an output size of the GRU size divided by 8 and the second layer output is equal to the hidden state size of the GRU.
These parameters were chosen following an exhaustive grid search, however, we found that any reasonable setting generally worked.

\subsection{The DeepAlign Algorithm}
In the context of processes, the sentences of words from above will become the cases of events from the event log.
By replacing the next word prediction RNNs with next event prediction RNNs in the BiBS algorithm we can apply it to event logs.
Instead of only predicting the next word, the RNNs will predict the next event, including the most likely event attributes.

Our goal is to utilize the two RNNs as the reference model for conformance checking and produce an alignment between log and the RNNs.
Alignments can be interpreted as a sequence of \textit{skip} (synchronous move), \textit{insertion} (model move), or \textit{deletion} (log move) operations.
The BiBS algorithm already covers the first two operations, but not the last.
To allow for deletions, we have to extend the BiBS algorithm.

Let $\overrightarrow{\text{RNN}}$ be the forward event prediction RNN and $\overleftarrow{\text{RNN}}$ be the backward RNN.
Let further $\text{RNN}(h, c)$ be the probability of case $c$ under $\text{RNN}$, initialized with the hidden state $h$.

\begin{figure}[t]
  \centering
  \includegraphics[width=1.0\linewidth]{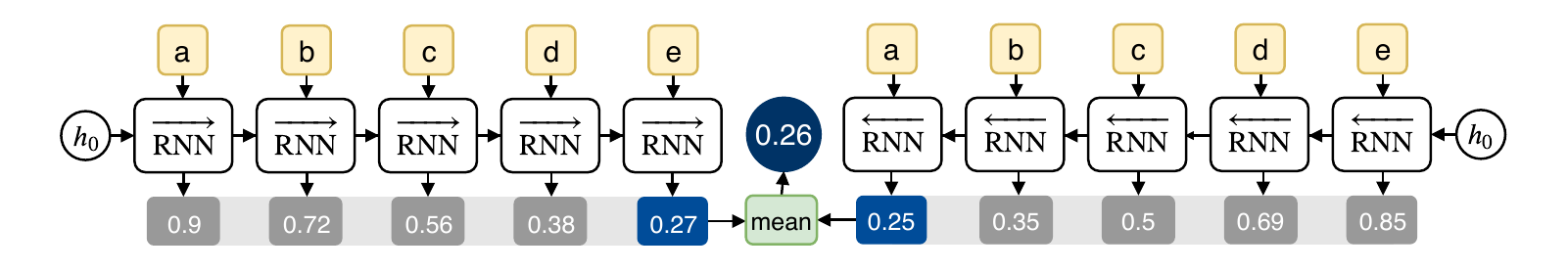}
  \caption{The probability of a case $c = \langle a,b,c,d,e \rangle$ is computed by the average probability of the case under both the forward and the backward RNN}
  \label{fig:indel-skip}
\end{figure}

The probability of a case $c$ under the two RNNs can be computed by
\begin{equation*}
  P(c) = \frac{1}{2} \left( \overrightarrow{\text{RNN}} \left( h_0, c \right) + \overleftarrow{\text{RNN}} \left( h_0, c \right) \right),
\end{equation*}
where $h_0$ is the output of the case attribute network.
If no case attributes are available, the initial state is set to zeros.
An example is shown in Fig.~\ref{fig:indel-skip}.

\begin{figure}[t]
  \centering
  \includegraphics[width=1.0\linewidth]{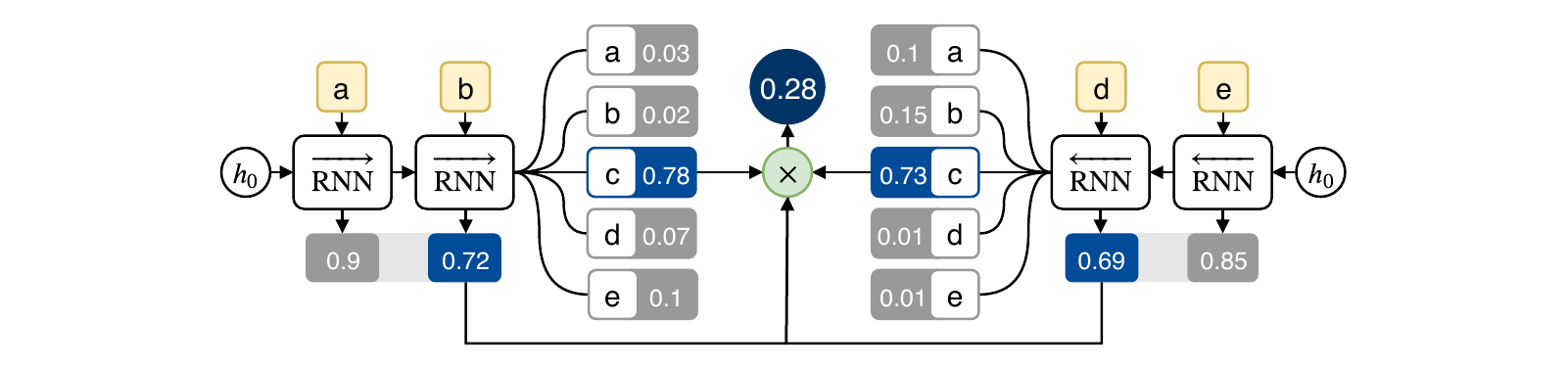}
  \caption{The probability of a case $c = \langle a,b,d,e \rangle$ after the insertion of an event $c$ after $b$ is computed by the joint probability $\langle a,b \rangle$ under the forward RNN, $\langle d,e \rangle$ under the backward RNN, and the probabilities of continuing the case with $c$ under both RNNs}
  \label{fig:indel-insert}
\end{figure}

For an insertion of an event $e$ at time $t$ in a case $c$, the probability under the two RNNs can be approximated by
\begin{align*}
  P_{\text{ins}}(c, e, t) = &~ \overrightarrow{\text{RNN}} \left( h_0, c_{[1:t]} \right) \cdot \overrightarrow{\text{RNN}} \left( \overrightarrow{h}_{t}, e \right) \\ 
  &\cdot \overleftarrow{\text{RNN}} \left( \overleftarrow{h}_{t+1}, e \right) \cdot \overleftarrow{\text{RNN}} \left( h_0, c_{[t+1:T]} \right),
\end{align*}
where $T$ is the total case length, $c_{[1:t]}$ is the index notation to retrieve all events from $c$ until time t, and $\overrightarrow{h}_t$ is the hidden state of $\overrightarrow{\text{RNN}}$ after reading $c_{[1:t]}$.
Similarly, $\overleftarrow{h}_{t+1}$ is the hidden state of $\overleftarrow{\text{RNN}}$ after reading $c_{[t+1:T]}$.
An example is shown in Fig.~\ref{fig:indel-insert}.

\begin{figure}[t]
  \centering
  \includegraphics[width=1.0\linewidth]{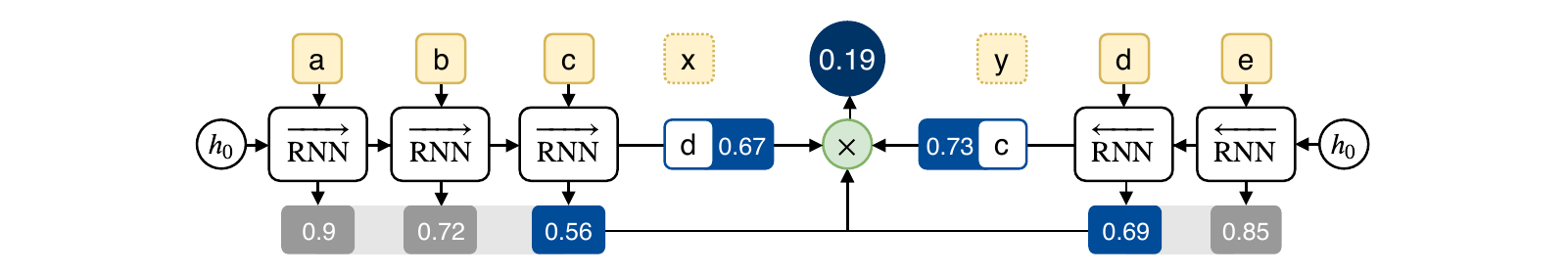}
  \caption{The probability of a case $c = \langle a, b, c, x, y, d, e \rangle$ after the deletion of $x$ and $y$ is computed by the joint probability of $\langle a,b,c \rangle$ under the forward RNN, $\langle d,e \rangle$ under the backward RNN, and the probabilities of continuing the case with $d$ and $c$ under the forward and backward RNN, respectively}
  \label{fig:indel-delete}
\end{figure}

The probability of deleting $n$ events at time $t$ in a case $c$ can be approximated by
\begin{align*}
  P_{\text{del}}(c, n, t) = &~ \overrightarrow{\text{RNN}} \left( h_0, c_{[1:t]} \right) \cdot \overrightarrow{\text{RNN}} \left( \overrightarrow{h}_{t}, c_{[t+n]} \right) \\
  &\cdot \overleftarrow{\text{RNN}} \left( \overleftarrow{h}_{t+n}, c_{[t]} \right) \cdot \overleftarrow{\text{RNN}} \left( h_0, c_{[t+n:T]} \right).
\end{align*}
An example is shown in Fig.~\ref{fig:indel-delete}.

Algorithm~\ref{alg:deepalign} shows the full DeepAlign process of aligning a case $c$ with the two RNNs.
The algorithm is initialized with an initial set of beams $B = \{c\}$, i.e., the original case.
For each possible operation, the probabilities are computed using the aforementioned equations, and the top-K beams are returned.
For simplicity, we assume that top-K always returns the updated cases according to the operations with the highest probability.
The number of events that can be deleted in one step can be controlled with the parameter $N$.
This is necessary because successively deleting single events does not necessarily generate higher probabilities than removing multiple events at once.

\begin{algorithm}
  \caption{DeepAlign algorithm}\label{alg:deepalign}
  \KwData{Given a set of beams $B$, maximum number of beams $K$, and a maximum deletion size $N$}
  \While{not converged}{
    $B' = \emptyset$\;
    \For{$b \in B$}{
      $B' = B' \cup P(b)$\;
      \For{$t = 1, ..., T$}{
        $B' = B' \cup \left\{ P_{\text{del}}(b, n, t)~|~n \in 1, ..., N \right\} \cup \left\{ P_{\text{ins}}(b, e, t)~|~e \in \mathcal{E} \right\}$
      }
    }
    $B = \text{top-K}\left( B' \right)$
  }
  \KwResult{$B$, the top-K beams after convergence}
\end{algorithm}

Algorithm~\ref{alg:deepalign} does not yet return alignments, but the top-K updated cases.
By keeping a history of the top-K operations (\textit{skip}, \textit{deletion}, and \textit{insertion}) in every iteration, we can obtain the alignment directly from the history of operations.
A \textit{deletion} corresponds to an empty move on the model, whereas an \textit{insertion} corresponds to an empty move in the log.

The top-K selection in Alg.\ref{alg:deepalign} will select the top K beams based on the probability under the RNN models.
In case of ties, we break the tie by choosing the beam with fewer empty moves (insertions and deletions).

\section{Experiments}\label{sec:experiments}
We evaluate the DeepAlign algorithm for the task of anomaly correction.
Given an event log consisting of normal and anomalous cases, an anomaly correction algorithm is expected to align each case in the event log with a correct activity sequence according to the process (without anomalies) that produced the event log.

\begin{figure}[t]
  \centering
  \includegraphics[width=1.0\linewidth]{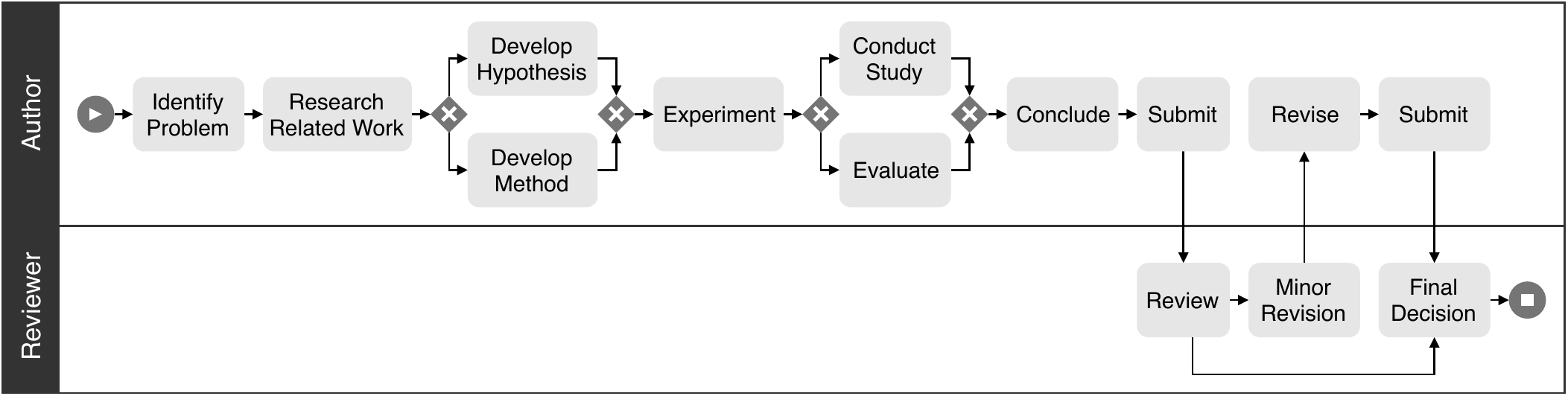}
  \caption{A simple paper submission process which is used as an example in the evaluation}
  \label{fig:process}
\end{figure}

We use a simple paper submission process as a running example throughout the remainder of this paper.
The process model in Fig.~\ref{fig:process} describes the creation of a scientific paper.
It includes the peer review process, which is executed by a reviewer, whereas the paper is written by an author.

To evaluate the accuracy of the corrections, we generated six random process models using PLG2~\cite{burattin2016plg2}.
The models vary in complexity with respect to the number of activities, breadth, and width.
Additionally, we use a handmade procurement process model called P2P as in~\cite{nolle2018binet}.

\begin{figure}[t]
  \centering
  \includegraphics[width=1.0\linewidth]{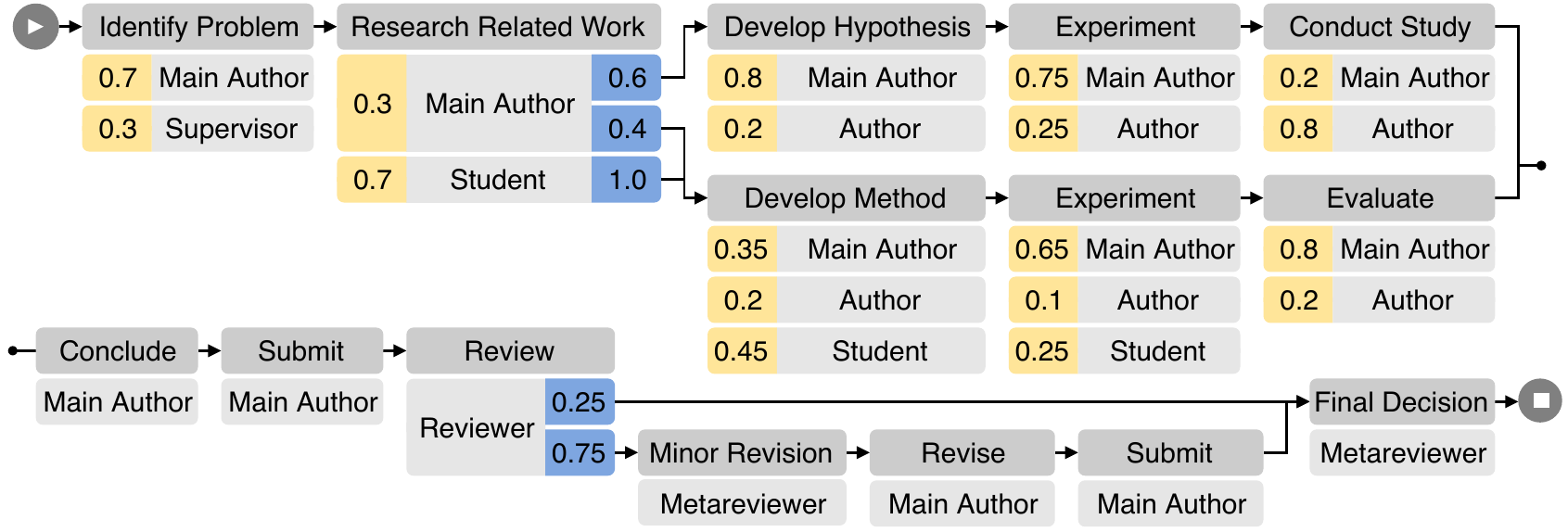}
  \caption{A likelihood graph with user attribute; 1.0 probabilities omitted for simplicity}
  \label{fig:likelihood}
\end{figure}

To generate event attributes, we create a likelihood graph~\cite{bohmer2016multi} from the process models which includes probabilities for event attributes connected to each step in the process.
This method has been proposed in~\cite{nolle2019binet}.
A likelihood graph for the paper process from Fig.~\ref{fig:process} is shown in Fig.~\ref{fig:likelihood}.

For each process step, the probability of the resource executing it is shown in yellow.
Depending on the resource, the probabilities of the next process steps are shown in blue.
Note that there is a long-term dependency between the steps \textit{Develop Hypothesis} and \textit{Conduct Study}, and, similarly, between \textit{Develop Method} and \textit{Evaluate}.
That is, \textit{Conduct Study} never eventually follows \textit{Develop Method}, and, likewise, \textit{Evaluate} never eventually follows \textit{Develop Hypothesis}.

We can generate event logs by using a random-walk through the likelihood graph, complying with the transition probabilities, and generating activities and attributes along the way.
In addition to the event attributes, we also generate case attributes, as well as, dependencies between the case attributes and the output probabilities in the likelihood graph.
For the paper process, we generate two case attributes, \textit{Decision} and \textit{Topic}.

If the topic is \textit{Theory}, this implies that \textit{Develop Hypothesis} will occur in a case, whereas if the topic is \textit{Engineering}, it implies \textit{Develop Method} will occur.
The decision can be \textit{Accept}, \textit{Weak Accept}, \textit{Borderline}, \textit{Weak Reject}, or \textit{Reject}.
For simplicity, we define that there will only be a \textit{Minor Revision} if the \textit{Decision} is either \textit{Accept} or \textit{Weak Accept}.
There will be no \textit{Minor Revision} otherwise.
We have generated an event log that follows these rules that we use as an example throughout the remainder of the paper.
The paper process was not used in the evaluation because of its simplicity.

For each of the 7 other process models, we generate 4 random event logs with varying numbers of event and case attributes.
Additionally, we introduce noise to the event logs by randomly applying one of 7 anomalies to a fixed percentage of the cases in the event log.
We generate datasets for noise levels between 10\% and 90\% with a step size of 10\% (9 in total).
We gather a ground truth dataset for the anomaly correction problem by retaining the original cases before alteration.
The 7 anomalies are defined as follows.
\begin{itemize}
  \item \textit{Skip}: A sequence of up to 2 necessary events has been skipped
  \item \textit{Insert}: Up to 2 random activities have been inserted
  \item \textit{Rework}: A sequence of up to 3 events has been executed a second time
  \item \textit{Early}: A sequence of up to 2 events has been executed too early, and hence is skipped later in the case
  \item \textit{Late}: A sequence of up to 2 events has been executed too late, and hence is skipped earlier in the case
  \item \textit{Attribute}: An incorrect attribute value has been set in up to 3 events
\end{itemize}

To analyze the impact of the case and event attributes, we evaluate four different implementations of DeepAlign: one that does not use any attributes (DeepAlign$\emptyset$), one that only uses case attributes (DeepAlignC), one that only uses event attributes (DeepAlignE), and one that uses both (DeepAlignCE).

Additionally, we evaluate baseline approaches that first discover a process model using a discovery algorithm and then calculate the alignments~\cite{adriansyah2013memory}.
We chose the Heuristics Miner~\cite{weijters2011flexible} and the Inductive Miner~\cite{leemans2014process} using the implementations of the PM4Py library~\cite{berti2019process}.
For completeness, we also evaluate the conformance checking algorithm using a perfect Reference Model, i.e., the one used to generate the event logs.

We run the DeepAlign algorithm for a maximum number of 10 iterations with a beam size of $K=5$ and a maximum deletion size of $N=3$, and consider the top-1 beam for the evaluation.
The Inductive Miner and the Heuristics Miner are used as implemented in PM4Py.
For the Heuristics Miner, we use a dependency threshold of $0.99$, and for the Inductive Miner, we use a noise threshold of $0.2$.

\section{Evaluation}\label{sec:Evaluation}
The overall results are shown in Tab.~\ref{tab:results}.
For each dataset we run the algorithms and evaluate the correction accuracy, that is, an alignment is regarded as correct if the model sequence is exactly equal to the ground truth sequence.
For correct alignments, we calculate the optimality of the alignment (i.e., if the number of empty moves is minimal).
For incorrect alignments, we calculate the distance from the ground truth sequence with Levenshtein's algorithm.
Accuracy is measured as the macro average $F_1$ score of normal ($F_1^N$) and anomalous ($F_1^A$) cases across all datasets and noise levels.
\begin{table}
  \centering
  \caption{Correction accuracy, average error  for incorrect alignments (based on the Levenshtein distance), and alignment optimality for correct alignments; best results are shown in bold typeface}
  \setlength{\tabcolsep}{3pt}
  \label{tab:results}
  \begin{tabular}{lcccrrrrr}
    \toprule
                         & CF     & CA     & EA     & $F_1^N$         & $F_1^A$         & $F_1$           & Error         & Optimal         \\
    \midrule
    Reference Model      & \cmark & -      & -      & 0.9011          & 0.9331          & 0.9171          & \textbf{1.46} & -               \\
    \midrule
    Heuristics Miner     & \cmark & -      & -      & 0.6678          & 0.6144          & 0.6411          & 3.33          & -               \\
    Inductive Miner      & \cmark & -      & -      & 0.6007          & 0.2438          & 0.4222          & 2.18          & -               \\
    \midrule
    DeepAlign$\emptyset$ & \cmark & -      & -      & 0.7950          & 0.8111          & 0.8030          & 2.52          & 99.8\%          \\
    DeepAlignC           & \cmark & \cmark & -      & 0.8918          & 0.9290          & 0.9104          & 2.41          & \textbf{99.9\%} \\
    DeepAlignE           & \cmark & -      & \cmark & 0.9261          & 0.9582          & 0.9421          & 1.65          & 86.9\%          \\
    DeepAlignCE          & \cmark & \cmark & \cmark & \textbf{0.9442} & \textbf{0.9702} & \textbf{0.9572} & 1.84          & 86.6\%          \\
    \bottomrule
  \end{tabular}
\end{table}

Interestingly, DeepAlignE, and DeepAlignCE both outperform the perfect Reference Model approach.
This is because the Reference Model does not contain any information about the case and event attributes.
The Heuristics Miner yields much better results in the anomaly correction task than the Inductive Miner, however, DeepAlign$\emptyset$ outperforms both, without relying on case or event attributes.

\begin{figure}[t]
  \centering
  \includegraphics[width=1.0\linewidth]{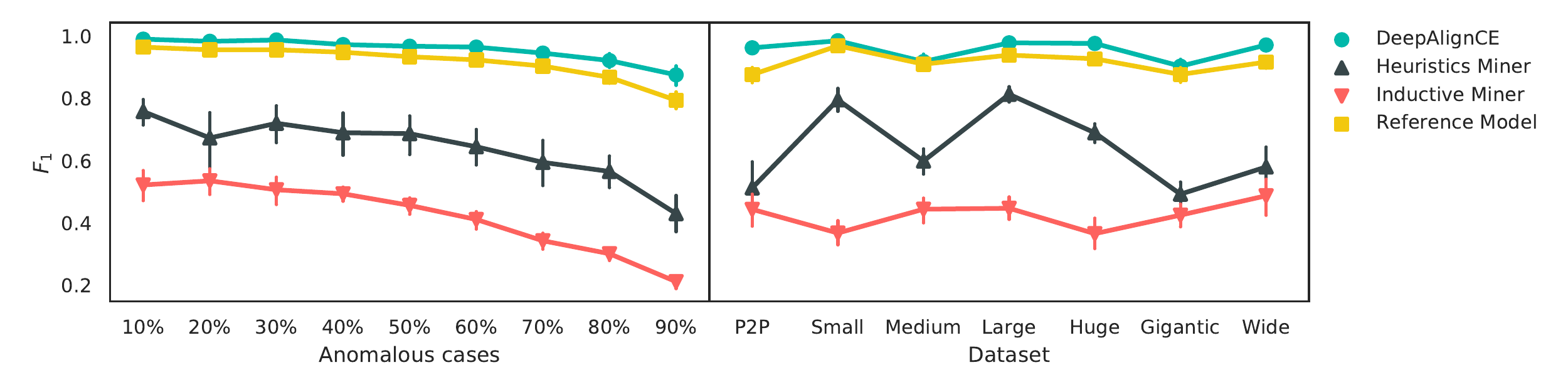}
  \caption{$F_1$ score for each algorithm per noise ratio (left) and per dataset (right); error bars indicate variance across all runs}
  \label{fig:details}
\end{figure}

Reference Model, Heuristics Miner, and Inductive Miner all produce optimal alignments because the alignment algorithm guarantees it.
The DeepAlign algorithm shows a significant drop in alignment optimality when including the event attributes.
The drop in optimality can be attributed to the fact that we always predict the top-1 attribute value for inserted events in the DeepAlign algorithm.
Furthermore, it might be connected to the attribute level anomalies that we introduced as part of the generation.
The best results are achieved when including both the case and the event attributes. 
Figure~\ref{fig:details} shows the $F_1$ score for each algorithm per noise level and per dataset.
DeepAlignCE always performs better than the Reference Model, and significantly better than the two mining approaches.

We want to finish the evaluation with examples from the paper dataset to illustrate the results of the DeepAlign algorithm. This is the resulting alignment for a case with a \textit{Skip} anomaly,
\begin{center}
  \setlength{\tabcolsep}{3pt}
  \begin{scriptsize}
    \begin{tabular}{|c|c|c|c|c|c|c|c|c}
      \makecell{Identify\\ Problem} & $\gg$ & $\gg$ & Experiment   & Evaluate & Conclude & Submit & Review & ... \\
      \hline
      \makecell{Identify\\ Problem} & \makecell{Research\\Related\\Work} & \makecell{Develop\\Method} & Experiment & Evaluate   & Conclude & Submit & Review & ...
    \end{tabular} 
  \end{scriptsize}
\end{center}
this is the result for a case with a \textit{Late} anomaly,
\begin{center}
  \setlength{\tabcolsep}{3pt}
  \begin{scriptsize}
    \begin{tabular}{|c|c|c|c|c|c|c|c|c|c}
      \makecell{Identify\\ Problem} & $\gg$ & $\gg$ & Experiment & \makecell{Research\\Related\\Work} & \makecell{Develop\\Method} & Evaluate & Conclude & Submit &... \\
      \hline
      \makecell{Identify\\ Problem} & \makecell{Research\\Related\\Work} & \makecell{Develop\\Method} & Experiment & $\gg$ & $\gg$ &  Evaluate & Conclude & Submit &...
    \end{tabular}
  \end{scriptsize}
\end{center}
and this is the result for a case with an \textit{Insert anomaly}.
\begin{center}
  \begin{scriptsize}
    \setlength{\tabcolsep}{3pt}
    \begin{tabular}{|c|c|c|c|c|c|c|c|c|c}
      \makecell{Identify\\ Problem} & \makecell{Research\\Related\\Work} & \makecell{Random\\activity\\10} & \makecell{Develop\\Method} & Experiment & Evaluate & Conclude & \makecell{Random\\activity\\12} & Submit & ... \\
      \hline
      \makecell{Identify\\ Problem} & \makecell{Research\\Related\\Work}  & $\gg$ & \makecell{Develop\\Method} & Experiment & Evaluate & Conclude & $\gg$ & Submit & ...
    \end{tabular} 
  \end{scriptsize}
\end{center}

The DeepAlign method can also be utilized to generate sequences from nothing, that is, to align the empty case with the most likely case according to the model.
Depending on the case attributes that are used to initialize the RNNs, the results will be different.

For $\text{Decision} = \text{Reject}$ and $\text{Topic} = \text{Engineering}$ the resulting sequence is $\langle$\,Identify Problem, Research Related Work, Develop Method, Experiment, Evaluate, Conclude, Submit, Review, Final Decision\,$\rangle$, whereas if we set $\text{Topic} = \text{Theory}$ the resulting sequence is $\langle$\,Identify Problem, Research Related Work, Develop Hypothesis, Experiment, Conduct Study, Conclude, Submit, Review, Final Decision\,$\rangle$.
The DeepAlign algorithm correctly generates a sequence including the \textit{Develop Method} and \textit{Develop Hypothesis} activities according to the setting of the \textit{Topic} case attribute.
It also does not generate the \textit{Minor Revision} activity because the \textit{Decision} is \textit{Reject}.
When setting $\text{Decision} = \text{Accept}$, DeepAlign will generate the sequence including the \textit{Minor Revision} branch.
A similar effect can be observed when altering the event attributes.

This demonstrates that the RNNs are indeed capable of learning the rules behind the decisions in the paper process (cf.~\cite{tax2018discovery}).
Although the paper dataset contains unambiguous dependencies between the case attributes and the resulting correct sequences, the overall results on the randomly generated datasets indicate that case and event attributes ought not to be neglected.

\section{Related Work}\label{sec:related_work}
Anomaly detection in business processes is frequently researched.
Many approaches exist that aim to detect anomalies in a noisy event log (i.e., an event log that contains anomalous cases).

Bezerra et\,al. have proposed multiple approaches utilizing discovery algorithms to mine a process model and then use conformance checking to infer the anomalies~\cite{bezerra2013algorithms}.
B\"ohmer et\,al. proposed a technique based on an extended likelihood graph that is utilizing event-level attributes to further enhance the detection~\cite{bohmer2016multi}.
The approach from~\cite{bohmer2016multi} requires a clean event log (i.e., no anomalies in the log), but it has been shown that the same technique can be applied to noisy logs as well~\cite{nolle2019binet}.
Recently, Pauwels et\,al. presented an approach based on Bayesian Networks~\cite{pauwels2019anomaly}.
Deep learning based approaches are presented in~\cite{nolle2018analyzing} and~\cite{nolle2019binet}.
However, none of these approaches can be utilized to correct an anomalous case or to produce an alignment.

Since Bezerra et\,al. presented their approach based on discovery algorithms in 2013, Mannhardt et\,al. have proposed both a data-aware discovery algorithm~\cite{mannhardt2015multi} and a data-aware conformance checking algorithm~\cite{mannhardt2016balanced}.
The conformance checking algorithm relies on a configurable cost function for alignments that must be manually defined to include the case and event attributes.
Our approach does not rely on a manual definition of the cost function, it traverses the search space based on learned probabilities instead.

Although alignments represent the current state-of-the-art in conformance checking~\cite{adriansyah2013memory}, they often pose a significant challenge because they are computationally expensive.
Van Dongen et\,al. address this issue in~\cite{dongen2017aligning}, compromising between computational complexity and quality of the alignments.
Very recently, Leemans et\,al. have presented a stochastic approach to conformance checking~\cite{leemans2019earth}, which can speed up the computation.

All of these approaches either rely on a non-data-aware discovery technique, require a manual effort to create a proper cost function, or they cannot generate alignments.
To the best of our knowledge, DeepAlign is the first fully autonomous anomaly correction method.

\section{Conclusion}\label{sec:conclusion}
We have demonstrated a novel approach to calculate alignments based on the DeepAlign algorithm.
When no reference model is available, two recurrent neural networks can be used to approximate the underlying process based on execution data, including case and event attributes.
The empirical results obtained in the experiments indicate that RNNs are indeed capable of modeling the behavior of a process solely based on an event log event if it contains anomalous behavior.

To the best of our knowledge, this is the first time that deep learning has been employed to calculate alignments in the field of process mining.
Although we evaluate DeepAlign in the context of anomaly correction, many other applications are conceivable.
For example, instead of training on a log that contains anomalies, a clean log could be used.
Furthermore, a clean log can be obtained from an existing reference model, and DeepAlign could be used to find alignments.
In other words, it might be possible to convert a manually created process model into a DeepAlign model.
A discovery algorithm based on DeepAlign is also imaginable since DeepAlign can also be utilized to generate sequences from scratch.
Depending on the case attributes the resulting predicted sequences will be different.
We think that this idea lends itself to further research.

We further believe that the DeepAlign algorithm could be employed to reduce the memory consumption of an alignment algorithm since the search space is efficiently pruned during the bidirectional beam search.
However, on the downside, DeepAlign does not guarantee optimal alignments.
This weakness can be addressed by employing an optimal alignment algorithm between the input sequence and the corrected sequence, albeit at the expense of efficiency.

In summary, DeepAlign is a novel and flexible approach with great application potential in many research areas within the field of process mining.

\section*{Acknowledgments}\label{sec:Acknowledgments}
This work is funded by the German Federal Ministry of Education and Research (BMBF) Software Campus project ``R2PA'' [01IS17050], Software Campus project ``KADet'' [01IS17050], and the research project ``KI.RPA'' [01IS18022D].

\bibliographystyle{splncs03}
\bibliography{references}

\end{document}